\documentclass[pdflatex,sn-mathphys]{sn-jnl}

\usepackage{multirow}
\usepackage{makecell}
\usepackage{booktabs}
\usepackage{xfrac}
\jyear{2022}%

\theoremstyle{thmstyleone}%
\theoremstyle{thmstyletwo}%

\theoremstyle{thmstylethree}%

\begin{document}

\title[A spatio-temporal network for video semantic segmentation in surgical videos]{A spatio-temporal network for video semantic segmentation in surgical videos}


\author*[1]{\fnm{Maria} \sur{Grammatikopoulou}}\email{maria.grammatikopoulou@medtronic.com}

\author[1]{\fnm{Ricardo} \sur{Sanchez-Matilla}}

\author[1]{\fnm{Felix} \sur{Bragman}}

\author[1]{\fnm{David} \sur{Owen}}

\author[1]{\fnm{Lucy} \sur{Culshaw}}

\author[1]{\fnm{Karen} \sur{Kerr}}

\author[1,2]{\fnm{Danail} \sur{Stoyanov}}

\author[1]{\fnm{Imanol} \sur{Luengo}}

\affil*[1]{\orgdiv{Medtronic Digital Surgery}, \orgaddress{\city{London}, \country{UK}}}

\affil[2]{\orgdiv{Wellcome/EPSRC Centre for Internventional and Surgical Sciences}, \orgname{University College London}, \orgaddress{\city{London}, \country{UK}}}


\abstract{Semantic segmentation in surgical videos has applications in intra-operative guidance, post-operative analytics and surgical education. Segmentation models need to provide accurate and consistent predictions since temporally inconsistent identification of anatomical structures can impair usability and hinder patient safety. Video information can alleviate these challenges; leading to reliable models suitable for clinical use. We propose a novel architecture for modelling temporal relationships in videos. Our model includes a spatio-temporal decoder to enable video semantic segmentation by improving temporal consistency across frames. An encoder processes individual frames whilst the decoder processes a temporal batch of adjacent frames. The decoder can be used on top of any segmentation encoder to improve temporal consistency. Model performance was evaluated on the CholecSeg8k dataset and a private dataset of robotic Partial Nephrectomy procedures. Segmentation performance was improved when the temporal decoder was applied across both. Our model also displayed improvements in temporal consistency. This work demonstrates an advance in video segmentation of surgical scenes with potential applications in surgical education and operating room guidance with a view to improve patient outcomes.
}

\keywords{Video segmentation, Semantic segmentation}



\maketitle

\section{Introduction}\label{sec:intro}
Video semantic segmentation has the potential to provide useful insights to surgeons by localising objects of interest in the surgical scene. Post-operative video processing can facilitate case review and improve training, while real-time guidance could support surgeons in uncertain scenarios through overlays. One method of extracting relevant information from surgical videos is scene segmentation, which provides per-pixel labelling of the scene. Anatomy localisation, facilitated by this technique, needs to be accurate and temporally consistent to provide surgeons with helpful guidance through model predictions. 

Scene segmentation in surgical videos has been studied extensively in the literature. State-of-the-art models often use single images as input to the model, producing inconsistent predictions, especially in sequences that contain images with ambiguous and partially occluded views. This could confuse surgeons and pose risks to the patient by providing contradictory information. This problem can be alleviated by using image sequences segmenting the scene, allowing the model to use temporal context within the video. By making predictions temporally consistent, the model is more reliable in challenging video sequences where the view can be partially occluded, or the anatomy of interest has not been fully exposed, in which case limited visual cues from single frames can cause models to flicker between plausible categorisations. 

In this paper, we propose a spatio-temporal model that uses features extracted from a series of consecutive frames by a single-frame encoder to provide temporally and spatially consistent predictions. We validate the proposed model using different single-frame encoders and against a state-of-the-art video segmentation model. Results are reported in two datasets, the publicly available semantic segmentation CholecSeg8k~\cite{hong2020CholecSeg8k} dataset, which includes images from laparoscopic cholecystectomy videos, and a private semantic segmentation dataset consisting of 137 Partial Nephrectomy (PN) procedures.

The first demonstration of the proposed temporal model is the detection of anatomy in PN videos, which is a novel application for segmentation models. In PN, it is important to correctly identify and expose the renal vein and artery in order to clamp the renal artery before excising the tumour from the kidney. Hence, we apply our temporal decoder to this dataset to demonstrate consistent and smooth prediction of the relevant anatomy during PN. The results show that the performance of semantic segmentation models improve when using the proposed temporal decoder while also increasing their temporal consistency. The contributions of this work are:
\begin{figure}[h!]
\centering
\includegraphics[width=0.9\textwidth]{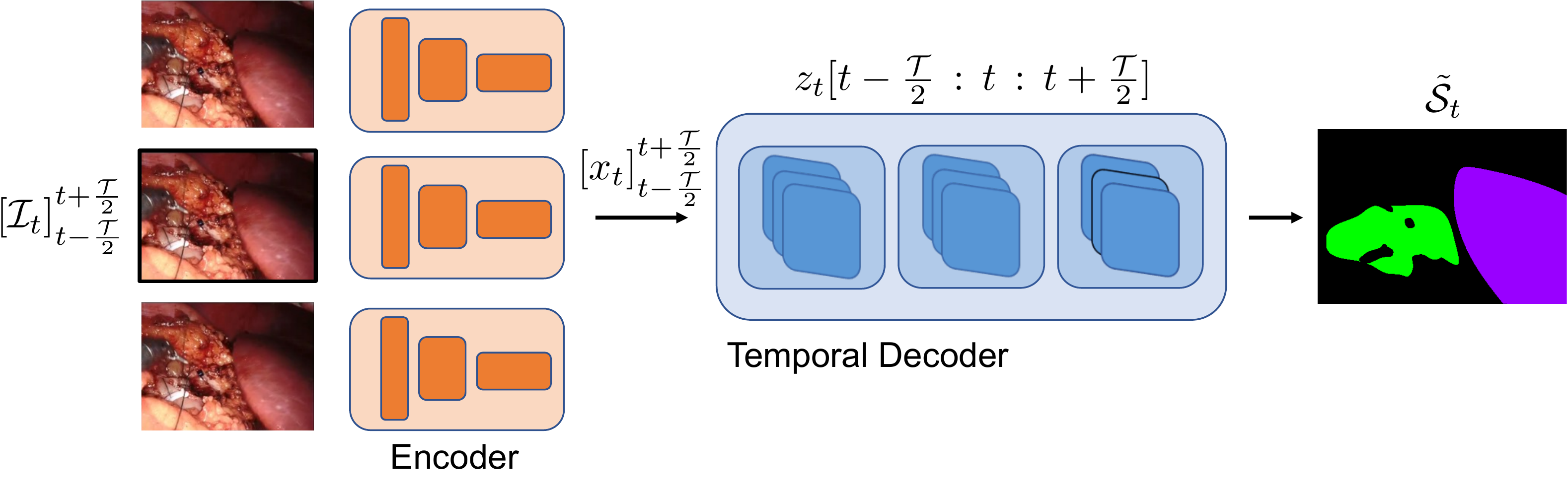}
\caption{Diagram of the proposed model. The model accepts a sequence of frames as an input based on a temporal window $\mathcal{T}$ and extracts features for each using a static encoder. They are subsequently passed to a temporal decoder which learns spatio-temporal representations and finally outputs a segmentation map for the central frame of the temporal batch.}\label{fig:model}
\end{figure}
\begin{itemize}
    \item A spatial and temporal convolutional model that can extend any single-frame segmentation architecture to leverage temporal context in semantic segmentation. Similar architectures have been used before for temporal classification \cite{Lea_2017_CVPR} tasks but have not been exploited for semantic segmentation.
    \item Quantitative investigation and benchmarking of the temporal consistency in two datasets and two different encoders. In addition to standard metrics reported for semantic segmentation, we evaluate the temporal consistency of the models using an optical flow-based metric inspired from \cite{varghese2020unsupervised}.
    \item Application of the proposed model to detection of anatomy in PN.
\end{itemize}
\section{Relevant work}\label{sec:literature}

A large number of semantic segmentation models, either convolutional-based~\cite{sun2019high} or transformer-based~\cite{liu2021swin}, rely on single images to identify objects in a scene. This can lead to spatially and temporally inconsistent predictions especially for ambiguous images for which the model needs temporal context. 

Previous work on video instance segmentation has used optical flow to track segmentation predictions~\cite{gonzalez2020isinet, zhao2020learning}. However, such methods are limited to using features between pairs of images and cannot leverage longest temporal context, while context aggregation also relies on the performance of the optical flow algorithm, which is computationally expensive. Transformer-based architectures have also been applied to tackle this problem, notably in~\cite{cheng2022masked} exploited mask-constrained cross-attention to learn temporal features between time-points in an architecture that performs both semantic and instance segmentation. Other methods have used a combination of 2D encoders and 3D Convolutional layers in the temporal decoder~\cite{puyal2020endoscopic} and Convolutional Long short-term Memory cells in the decoder~\cite{wang2021noisy}. Alternative approaches also include the enforcement of temporal consistency through a loss function during training~\cite{liu2020efficient} or through architectures that include high and low frame rate model branches to combine temporal context from different parts of the video~\cite{jain2019accel}. 
\begin{figure}[h!]
\centering
\includegraphics[width=0.9\textwidth]{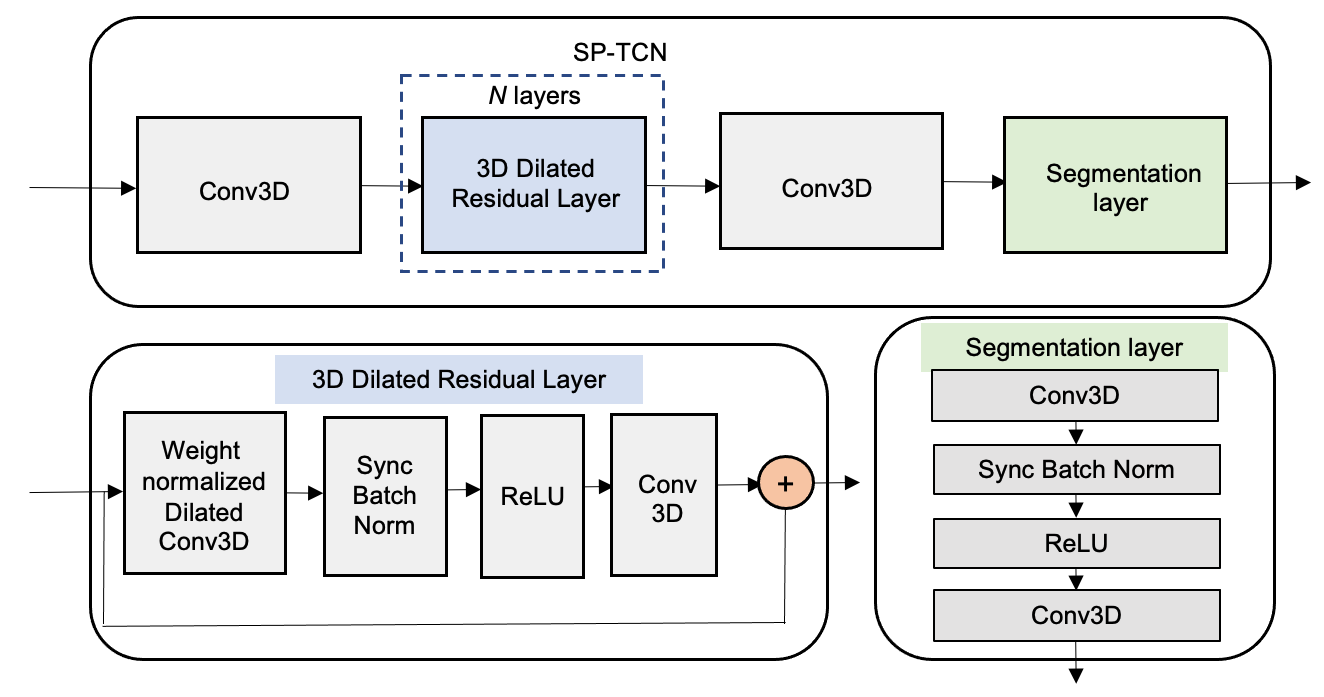}
\caption{Outline of the architecture of the proposed temporal decoder (SP-TCN).}\label{fig:model-blocks}
\end{figure}
Temporal modelling has been investigated thoroughly in action recognition~\cite{farha2019ms, hou2018spatial}. Temporal Convolutional Networks (TCNs) have been a popular architecture as they can provide a large receptive field without resulting on prohibitively large models and thus can operate on longer temporal windows. Variants of this model have been proposed, such as extending them to stochastic modelling~\cite{aksan2019stcn}; however, their benefits have not yet been employed to video semantic segmentation. Therefore, these benefits have been an inspiration to adapt this architecture to video semantic segmentation, which is by definition a computationally heavy task with one of the main challenges being the exploitation of longer temporal context.

\section{Methods}\label{sec:model}

%
Let $\mathcal{I}_t \in \{0,255\}^{W,H,C}$ be an RGB frame at time $t$ with width $W$, height $H$, and $C=3$ colour channels. Let $\mathcal{S}_t \in \{0,C\}^{W,H}$ be the corresponding pixel-wise segmentation annotation at time $t$ with $C$ semantic classes.
Let $E(\cdot)$ be an encoder that extracts frame representations for each frame individually as $E(\cdot) : \mathcal{I}_t \rightarrow \mathbf{x}_t$, where $\mathbf{x}_t$ is a spatial feature representation of the frame $\mathcal{I}_t$ at time $t$. We developed a temporal decoder $D(\cdot): [ \textbf{z}_t ]_{t=t-\sfrac{\mathcal{T}}{2}}^{t+\sfrac{\mathcal{T}}{2}} \rightarrow [\tilde{\mathcal{S}}_t]_{t=t-\sfrac{\mathcal{T}}{2}}^{t+\sfrac{\mathcal{T}}{2}}$ that processes a temporal batch of features $[\textbf{z}_t]_{t=t-\sfrac{\mathcal{T}}{2}}^{t+\sfrac{\mathcal{T}}{2}}$ centered at time $t$ within a temporal window of $\mathcal{T}$ frames. The result is a spatio-temporal decoder $\Lambda(\cdot): [\mathcal{I}_t]_{t=t-\sfrac{\mathcal{T}}{2}}^{t+\sfrac{\mathcal{T}}{2}} \rightarrow [\tilde{\mathcal{S}}_t]_{t=t-\sfrac{\mathcal{T}}{2}}^{t+\sfrac{\mathcal{T}}{2}}$ which predicts temporally consistent and accurate segmentation maps $[ \tilde{\mathcal{S}}_t]_{t=t-\sfrac{\mathcal{T}}{2}}^{t+\sfrac{\mathcal{T}}{2}}$ .


\begin{figure}[h!]
\centering
\includegraphics[width=0.8\textwidth]{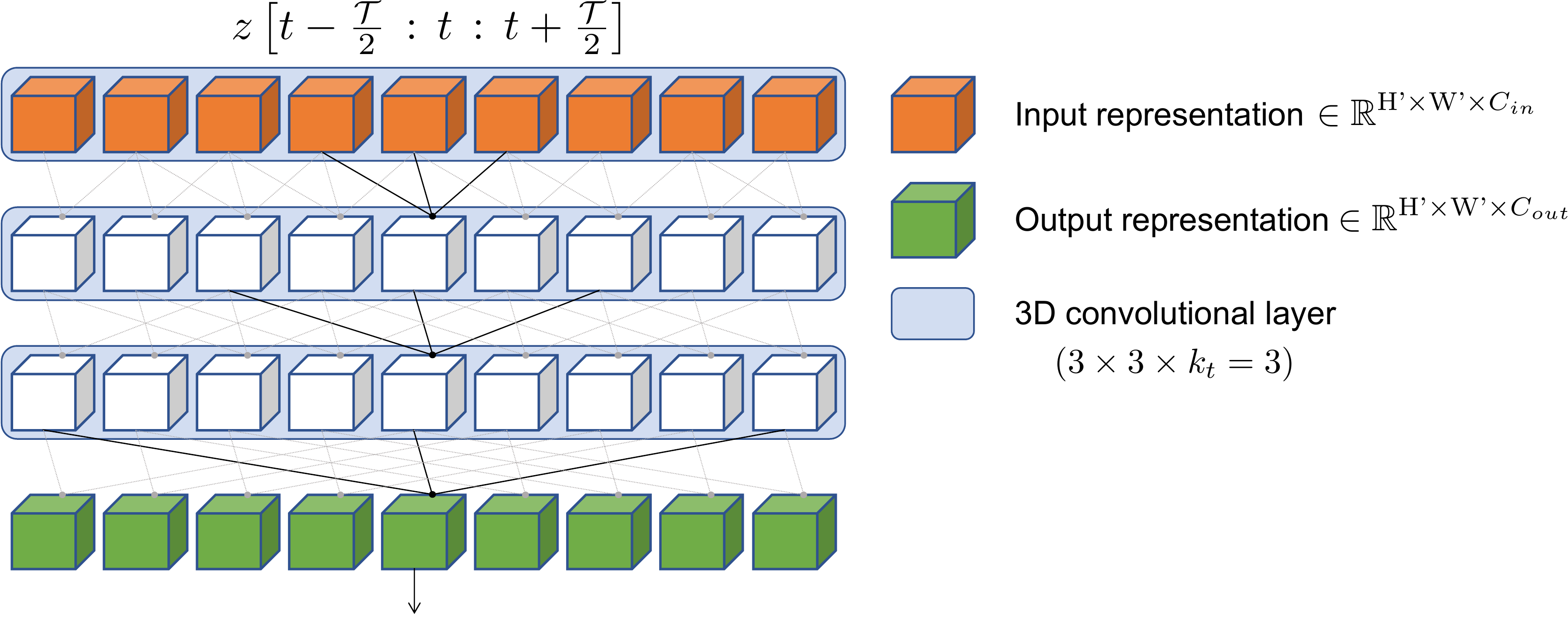}
\caption{Schematic of the increase in receptive field caused by repeated dilated 3D convolutions with $3$ successive layers using $k_t=3$ and $\mathcal{T}=8$. The exponential increase in the dilation factor facilitates a large temporal receptive field for each frame. Both inputs and outputs of the dilated TCN block are temporal representations $\mathbf{z}$.}\label{fig:dilation-blocks}
\end{figure}
\subsection{Spatial Temporal Convolutional Network (SP-TCN)}
The proposed decoder takes as input a temporal batch of static frame representations $[ \textbf{z}_t ]_{t=t-\sfrac{\mathcal{T}}{2}}^{t+\sfrac{\mathcal{T}}{2}}=E([ \textbf{x} ]_{t-\sfrac{\mathcal{T}}{2}}^{t+\sfrac{\mathcal{T}}{2}})$ from the encoder, centered at the image $\mathcal{I}_{t}$ where $\mathcal{T}$ is the temporal window of the model. The SP-TCN consists of three main building blocks: two 3D Convolutional blocks, $N$ 3D dilated residual layers and a segmentation layer. In Figure~\ref{fig:model-blocks}, we illustrate how dilation facilitates an exponential increase in the temporal receptive field in successive dilated 3D convolutions. Each convolutional layer consists of kernels of size $\left(3 \times 3 \times k_t\right)$ where $k_t$ determines the time kernel dimension. The convolutions are acausal, which process both past and future information. A representation $z_{t}$ consequently receives context from both $z_{t-k_t/2}$ and $z_{t+k_t/2}$. The 3D convolutional blocks both preceding and succeeding the $N$ 3D dilated residual layers are only composed of a single 3D convolutional layer. Conversely, each 3D dilated residual layer is composed of a dilated 3D convolution, followed by weight normalisation, batch normalisation, ReLU activation and a final 3D convolution layer. 

\textbf{3D Dilated Residual layers:} The dilated residual layers contribute towards a larger receptive field without increasing prohibitively increasing the depth of the network. The dilation factor  $d_i$ the $i_{th}$ dilated residual layer depends on the number of layers $N$ and is equal to $d_i = 2^{i}, \, \, \text{for}\, i=[0,\dots,N-1]$ where $i=0$ is the first layer. The full architecture of the dilated layers and the segmentation layer are shown in Figures \ref{fig:model-blocks} and \ref{fig:dilation-blocks}.

\section{Experimental validation}\label{sec:experiments}
The proposed model is benchmarked using two state-of-the-art encoders, the convolution-based light-weight version of HRNetv2~\cite{sun2019high} and Swin transformer~\cite{liu2021swin} to demonstrate how the temporal decoder improves state-of-the-art single-frame segmentation models of different size. In addition, we compare against the state-of-the-art video segmentation model Mask2Former~\cite{cheng2022masked}.

\subsection{Datasets}\label{sec:datasets}
The proposed model is benchmarked with two datasets, a private dataset consisting of images from PN procedures and the publicly available CholecSeg8k dataset which consists of images taken from a subset of laparoscopic cholecystectomy procedures~\cite{hong2020CholecSeg8k}.

\textbf{Partial Nephrectomy.}
The private PN dataset consists of 53,000 images from 137 procedures annotated with segmentation masks for the kidney, liver, renal vein, and renal artery. The images are annotated at 1 or 10 frames per second (fps) frames with sequence duration of 10 or 15 seconds. The images were labelled by trained non-medical experts under the supervision of an anatomy specialist, using annotation guidelines validated by surgeons. \\
\begin{figure}[h]
\centering
\includegraphics[width=0.9\textwidth]{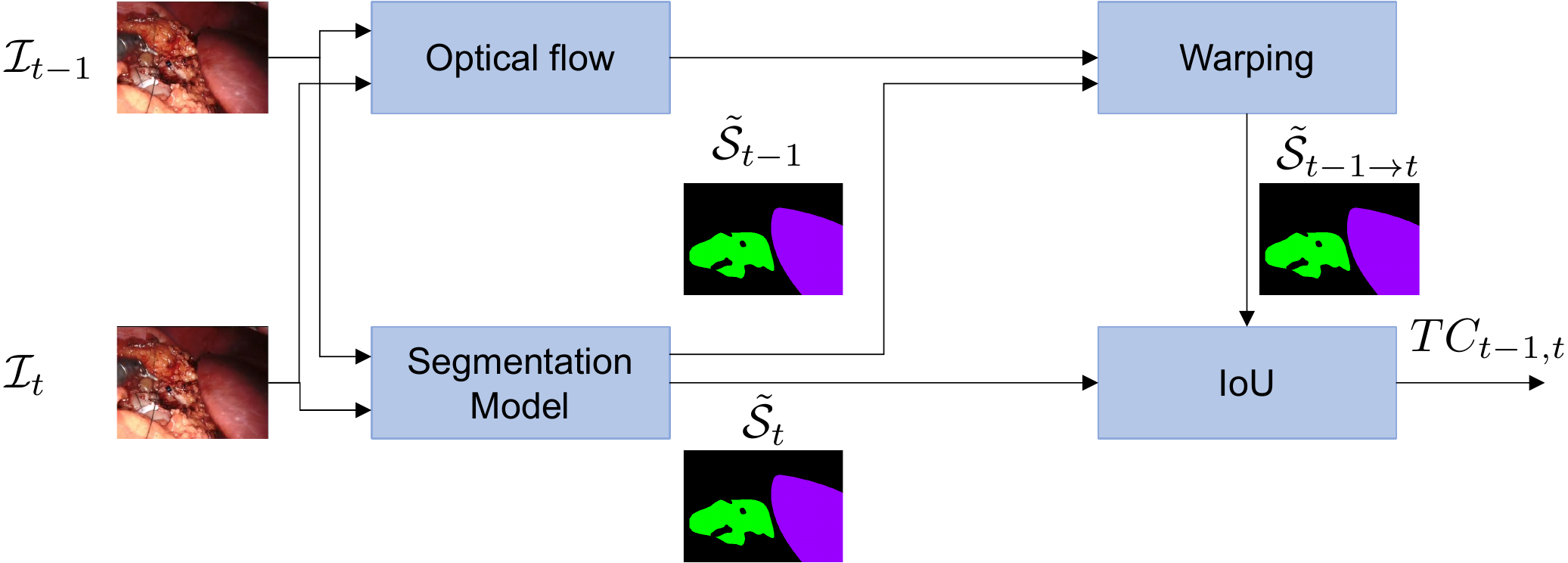}
\caption{Temporal consistency (TC) metric is calculated between a pair of consecutive frames $\mathcal{I}_{t-1}$ and $\mathcal{I}_{t}$. The frames are given as input to a pre-trained optical flow algorithm and the segmentation model under evaluation. The optical flow prediction warps $\tilde{\mathcal{S}}_{t-1}$ from $t-1$ to $t$, obtaining $\tilde{\mathcal{S}}_{t-1 \rightarrow t}$. The TC metric is calculated as the IoU between $\tilde{\mathcal{S}}_{t-1 \rightarrow t}$ and $\tilde{\mathcal{S}}_{t}$.}
\label{fig:tc-metric}
\end{figure}
\textbf{CholecSeg8k.}
The public CholecSeg8k dataset consists of 8,080 images from 17 videos of the Cholec80 dataset annotated at 25fps\cite{hong2020CholecSeg8k}. Images are annotated with segmentation masks containing 13 classes (background, abdominal wall, liver, gastrointestinal tract, fat, grasper, connective tissue, blood, cystic duct, l-hook electrocautery, gallbladder, hepatic vein and liver ligament).

\subsection{Metrics}
We assess the segmentation performance using the Intersection over Union (IoU) metric, and the temporal consistency of the model predictions using the Temporal Consistency (TC) metric..
The IoU is computed per each class and image, $IoU_t^c(S_t^c,\tilde{S}_t^c)=\frac{S_t^c \cap \tilde{S}_t^c}{S_t^c \cup \tilde{S}_t^c}$, where $S_t^c$ is the annotation, $\tilde{S}_t^c$ the model estimation for class $c$ on image at time $t$, $\cap$ is the intersection operator, and $\cup$ is the union operator. The IoU per class is computed as the mean across images $IoU^c = \frac{1}{\mathcal{T}} \sum_{t=1}^{\mathcal{T}}{IoU_t^c}$, where $T$ is the total number of images. We use the mean Intersection over Union (mIoU) across classes to report a single number, computed as $mIoU = \frac{1}{C} \sum_{c=1}^{C}{IoU^c}$ where $C$ is the total number of classes.
The TC metric, based on~\cite{varghese2020unsupervised}, is calculated as $TC_{t-1,t} = IoU(\tilde{S}_t, \tilde{S}_{t-1 \rightarrow{t}})$, where $\tilde{S}_{t-1 \rightarrow{t}}$ is the warped prediction from time $t-1$ to time $t$. We use RAFT~\cite{teed2020raft} as optical flow algorithm, pre-trained on Sintel dataset. Figure~\ref{fig:tc-metric} shows a visual representation of the TC metric calculation. 

\subsection{Experimental setup}
The temporal decoders were trained using $N=4$ dilated residual layers and feature size $128$ for each layer. We used the Adam optimizer, 1cycle learning rate scheduling and balanced sampling of classes with 500 samples during training. We chose a value of $k_t=3$ for the spatio-temporal convolutions. The model outputs a temporal batch of segmentations. However, we only backpropagate the loss on the central frame $\mathcal{I}_t$ of the temporal batch $[\mathcal{I}_{t}]_{t=t-\mathcal{T}/2}^{t+\mathcal{T}/2}$. The model was trained with a Cross Entropy loss. All models were trained for 100 epochs. The encoders trained in 12 hours and the decoders trained in 24 hours approximately. Models were implemented with PyTorch 1.10.

We used horizontal flipping as augmentation during training. For PN, we used 85\% of the videos for training, 5\% for validation, and 10\% for testing. For CholecSeg8k, since the dataset is small, we used 75\% of the videos for training and 25\% of the videos for testing (videos 12, 20, 48 and 55). The test set in CholecSeg8k was chosen to ensure that all classes had sufficient instances in the training set. For PN, the model weights for testing were selected based on the lowest validation loss. For CholecSeg8k, we used the weights in the last epoch due to the lack of validation set. We trained with a temporal window of $\mathcal{T}=14$ for PN and $\mathcal{T}=10$ for CholecSeg8k. 

Mask2Former was trained using pairs of frames selected within the same respective window sizes. For the PN dataset, we used unlabelled frames so that the windows include frames only at 10 fps and not a combination of frames of frame rates to facilitate model learning. For CholecSeg8k, we used the frames provided in the dataset only as they provide very dense temporal context with frames at 25 fps. All models were trained on 2 DGX A100 GPUs.

\subsection{Results}

We present quantitative and qualitative results for all models and datasets. Table \ref{tab:metrics-summary} summarises the mean IoU and mean TC for all models and all datasets. Results indicate that the segmentation performance improves  when using the proposed temporal decoder for both datasets in comparison to single-frame models. In particular, a 1.04\% to 1.3\% increase of the mean IoU is reported for PN with the use of the temporal decoder. Similarly, a 0.960\% to 4.27\% increase of the mean IoU is reported when using the SP-TCN with single-frame encoders compared to the single-frame model for CholecSeg8k. Similarly, results indicate that the temporal consistency improves with an increase of 6.29\%-7.23\% in PN, and an increase of 2.56-3.20\% in CholecSeg8k dataset. In both datasets,  the best performing combination is the Swin base encoder + SP-TCN.

Per-class metrics are presented in Table \ref{tab:pn-iou-per-class}, \ref{tab:pn-tc-per-class} for PN and for Cholecseg8k in Tables \ref{tab:ch-iou-per-class} and \ref{tab:ch-tc-per-class}. Results show that \textit{kidney} is the class that obtains the most consistent improvement across all combinations. Similar results are observed for CholecSeg8k as well, with Mask2Former giving similar performance to the best peforming Swin base + SP-TCN. It is worth noting that the absolute numbers for TC are higher for CholecSeg8k as the time interval between images is shorter than in the PN dataset (25 fps compared to 1 and 10 respectively), hence less motion is observed between frames and therefore there is higher overlap between predictions in subsequent frames.
\begin{figure}[h!]
\centering
\includegraphics[width=\textwidth]{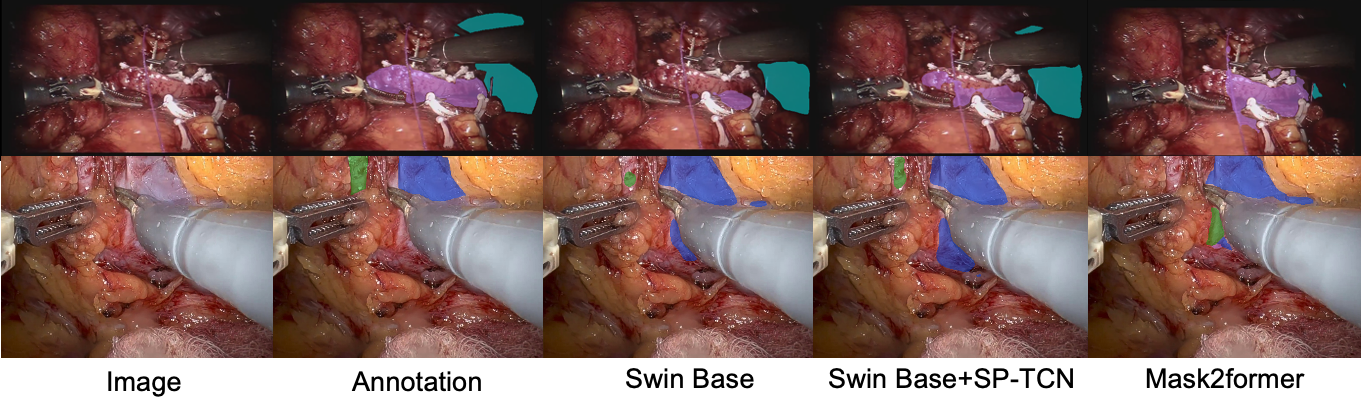}
\caption{Example predictions for the PN dataset. Figures show example segmentations for kidney (pink), liver (cyan), renal vein (blue) and renal artery (green).}
\label{fig:predictions-pn}
\end{figure}
\begin{figure}[h!]
\centering
\includegraphics[width=\textwidth]{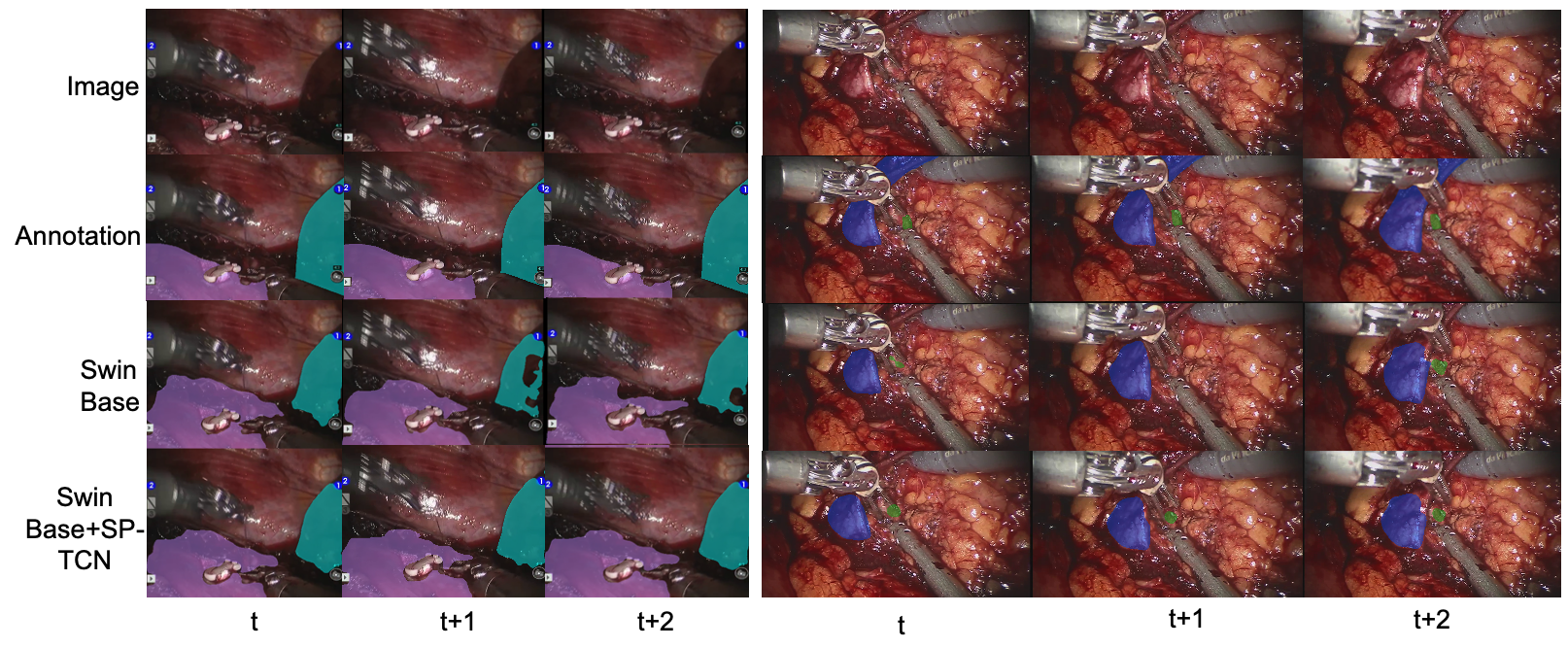}
\caption{Example predictions for kidney (pink), liver (cyan), renal vein (blue) and renal artery (green) for two sequences of three images from the PN dataset. Top row: images at three different timestamps of 0.1 seconds apart, second row: annotation, third row: Swin base, fourth row: Swin base + SP-TCN.}\label{fig:predictions-pn-seq}
\end{figure}

Example predictions for the PN dataset are shown in Figure~\ref{fig:predictions-pn}. These examples show that segmentation predictions are more temporally and spatially consistent in these sequences. For instance, the borders for the \textit{kidney} flicker less and the \textit{liver} segmentation does not miss any part of the anatomy between frames (left sequence). In addition, the temporal decoder recovers missed predictions by the single-frame model within the image sequence for the renal artery (right sequence).

The model presents some limitations which are the following: the temporal decoder may not be able to recover missing information if the features extracted by the encoder within the given temporal window do not include sufficient information. For example, if an anatomical structure is completely missed by the encoder in the full temporal window used by the decoder, the temporal decoder may fail to recover it. In addition, the images contained in the temporal window need to be of consistent time spacing since it does not learn sufficiently when used with images that come from varying time intervals. The temporal decoder also performs better when the images are within short time steps.

\begin{figure}[h]
\centering
\includegraphics[width=\textwidth]{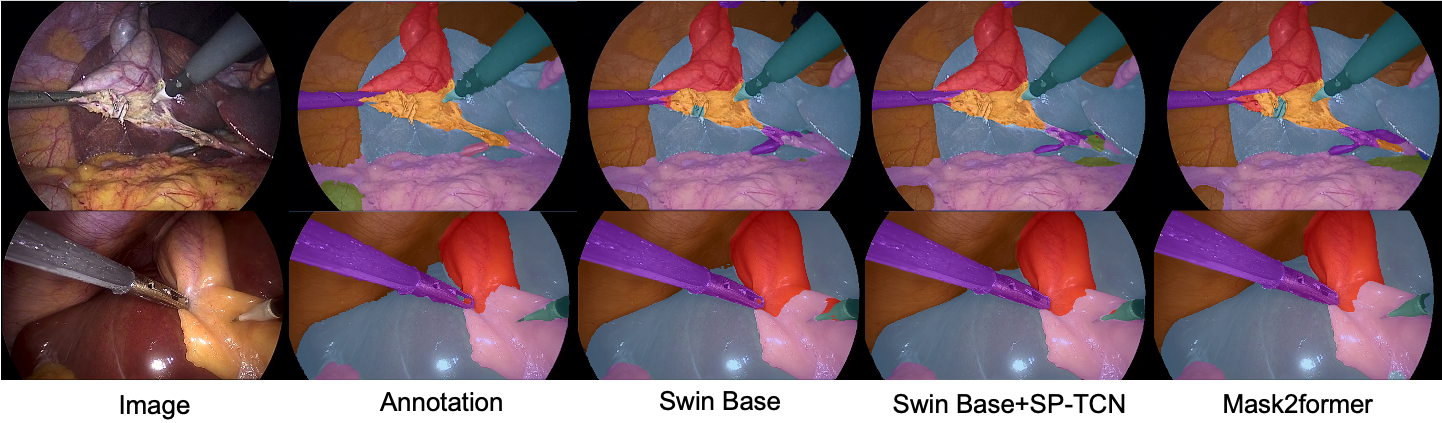}
\caption{Example predictions for the CholecSeg8k dataset for two sequences of three images. Top row: Images at three different timestamps of 0.04 seconds apart, second row: Annotation, third row: Swin base, fourth row: Swin base + SP-TCN.}\label{fig:predictions-cholecseg}
\end{figure}

\begin{figure}[h]
\centering
\includegraphics[width=\textwidth]{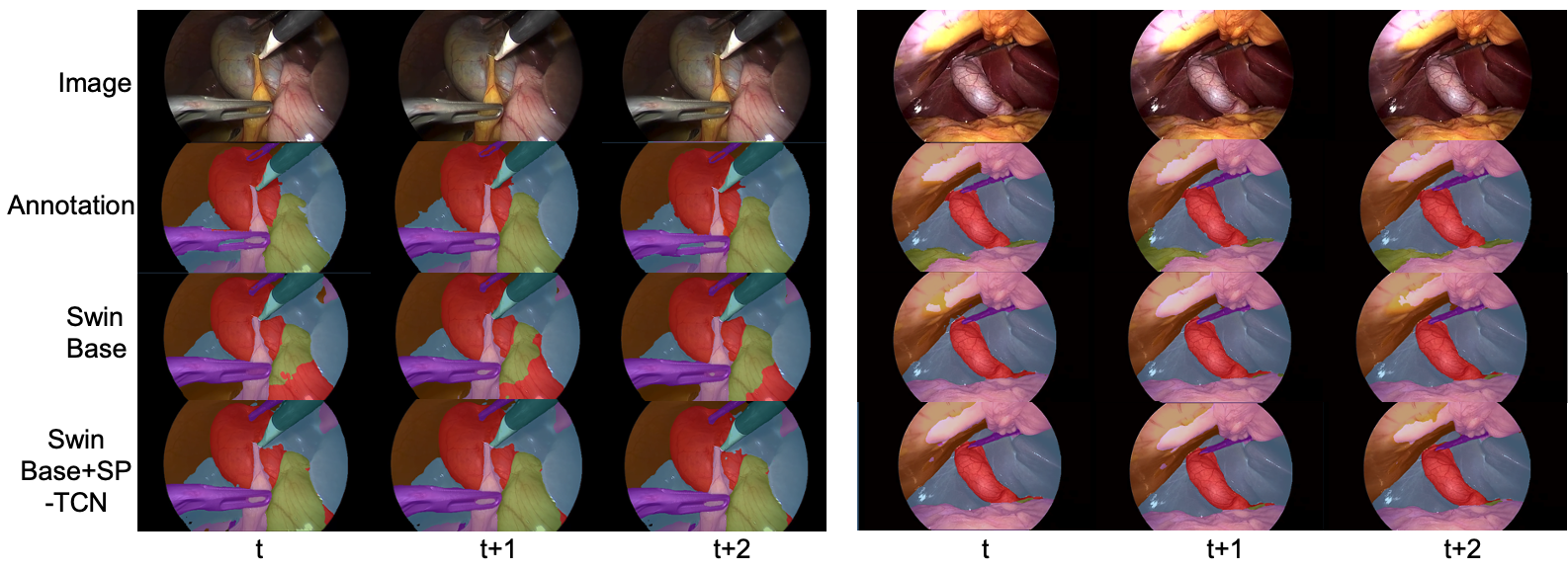}
\caption{Example predictions for the CholecSeg8k dataset for two sequences of three images. Top row: Images at three different timestamps of 0.04 seconds apart, second row: Annotation, third row: Swin base, fourth row: Swin base + SP-TCN.}\label{fig:predictions-cholecseg-seq}
\end{figure}
\begin{table}[h]
    \centering
    \caption{Mean IoU and Mean TC for PN and CholecSeg8k}\label{tab:metrics-summary}
    \begin{tabular}{lcccc}
    \specialrule{1.2pt}{0.2pt}{1pt}
    \textbf{Model} & \multicolumn{2}{c}{\textbf{PN}} & \multicolumn{2}{c}{\textbf{CholecSeg8k}} \\ 
    \cmidrule(lr){2-3}
    \cmidrule(lr){4-5}
     & Mean IoU & Mean TC & Mean IoU & Mean TC  \\
    \specialrule{1.2pt}{0.2pt}{1pt}
        Mask2Former     & 0.5424            & 0.4910 & 0.6910 & 0.8659  \\
        \hline
        HRNet32         & 0.5680            & 0.4570 & 0.6110 & 0.8368 \\
        HRNet32 + SP-TCN   & \textbf{0.5810}   & \textbf{0.5199} & \textbf{0.6537} & \textbf{0.8624} \\
        Difference (\%) & \cellcolor{green!20} 1.30 & \cellcolor{green!20} 6.29 & \cellcolor{green!20} 4.27 & \cellcolor{green!20} 2.56 \\
        \hline
        Swin base       & 0.6187 & 0.4890   & 0.6842 & 0.8406  \\
        Swin base +  SP-TCN & \textbf{0.6291}   & \textbf{0.5613} & \textbf{0.6938} & \textbf{0.8726} \\
        Difference (\%)& \cellcolor{green!20} 1.04 & \cellcolor{green!20} 7.23 & \cellcolor{green!20} 0.960 & \cellcolor{green!20} 3.20 \\
    \specialrule{1.2pt}{0.2pt}{1pt}
    \end{tabular}
\end{table}

\begin{table}[h]
    \centering
    \caption{Per-class IoU for PN}\label{tab:pn-iou-per-class}
    \resizebox{\textwidth}{!}{%
    \begin{tabular}{lc|cc|cc}
    \specialrule{1.2pt}{0.2pt}{1pt}
        \textbf{Class}  & \textbf{Mask2Former} & \textbf{HRNet32} & \multicolumn{1}{c}{\begin{tabular}[c]{@{}c@{}}\textbf{HRNet32} \\ \textbf{+SP-TCN}\end{tabular}} & \textbf{Swin Base} & \multicolumn{1}{c}{\begin{tabular}[c]{@{}c@{}}\textbf{Swin base} \\ \textbf{+SP-TCN}\end{tabular}} \\
        \specialrule{1.2pt}{0.2pt}{1pt}
            Background      &  0.9080  & 0.9140 & 0.9230 &	0.9270  &	0.9290  \\
            Kidney          &  0.5900  & 0.6180 & 0.6412 & 	0.6664  & 	0.6789  \\
            Liver           &  0.5265  & 0.5481 & 0.6186 & 	0.7120  &   0.7108  \\
            Renal artery    &  0.3345  & 0.3958 & 0.3900 & 	0.3855  & 	0.3966 \\
            Renal vein      &  0.3520  & 0.3672 & 0.3350 & 	0.4015  & 	0.4300  \\
            \hline
            Mean            &  0.5424  & 0.5680 & \textbf{0.5810} &  0.6187  &   \textbf{0.6291}  \\
    \specialrule{1.2pt}{0.2pt}{1pt}
    \end{tabular}
    }
\end{table}

\begin{table}[h]
    \centering
    \caption{Per-class TC for PN}\label{tab:pn-tc-per-class}
    \resizebox{\textwidth}{!}{%
    \begin{tabular}{lc|cc|cc}
        \specialrule{1.2pt}{0.2pt}{1pt}
        \textbf{Class} & \textbf{Mask2Former} & \textbf{HRNet32} & \multicolumn{1}{c}{\begin{tabular}[c]{@{}c@{}}\textbf{HRNet32} \\ \textbf{+SP-TCN}\end{tabular}} & \textbf{Swin Base} & \multicolumn{1}{c}{\begin{tabular}[c]{@{}c@{}}\textbf{Swin base} \\ \textbf{+SP-TCN}\end{tabular}}  \\
        \specialrule{1.2pt}{0.2pt}{1pt}
            Background      &  0.9346 & 0.9340 &  0.9443  & 	0.9362 & 	0.9455   \\
            Kidney          &  0.5950 & 0.5253 &  0.5661  & 	0.5466 & 	0.6014   \\
            Liver           &  0.4024 & 0.3499 &  0.4542  & 	0.3810 & 	0.4860   \\
            Renal artery    &  0.2220 & 0.2628 &  0.3497  & 	0.3188 & 	0.3912   \\
            Renal vein      &  0.3200 & 0.2143 &  0.2850  & 	0.2628 & 	0.3823   \\
            \hline
            Mean            &  0.4910 & 0.4570 &  \textbf{0.5190}  &   0.4890    &    \textbf{0.5613}   \\
        \specialrule{1.2pt}{0.2pt}{1pt}
    \end{tabular}
    }
\end{table}

\begin{table}[h]
    \centering
    \caption{Per-class IoU for CholecSeg8k}\label{tab:ch-iou-per-class}
    \resizebox{\textwidth}{!}{%
    \begin{tabular}{lc|cc|cc}
    \specialrule{1.2pt}{0.2pt}{1pt}
    \textbf{Class} & \textbf{Mask2Former} & \textbf{HRNet32} & \multicolumn{1}{c}{\begin{tabular}[c]{@{}c@{}}\textbf{HRNet32} \\ \textbf{+SP-TCN}\end{tabular}} & \textbf{Swin Base} & \multicolumn{1}{c}{\begin{tabular}[c]{@{}c@{}}\textbf{Swin base} \\ \textbf{+SP-TCN}\end{tabular}}  \\
    \specialrule{1.2pt}{0.2pt}{1pt}
        Background              & 	0.9785 & 0.9757    & 	0.9736 & 	0.9731 & 	0.9740  \\
        Abdominal wall          & 	0.6925 & 0.7329    & 	0.7444 & 	0.7848 & 	0.7414  \\
        Connective tissue       & 	0.2484 & 0.3490    & 	0.3537 & 	0.2510 & 	0.3120  \\
        Fat                     & 	0.8417 & 0.8350    & 	0.7608 & 	0.8408 & 	0.8415  \\
        Gallbladder             & 	0.6520 & 0.5599    & 	0.5651 & 	0.6040 & 	0.6109  \\
        Gastrointestinal tract  & 	0.5230 & 0.1930    & 	0.4490 & 	0.4935 & 	0.5714  \\
        Grasper                 & 	0.7375 & 0.5589    & 	0.6429 & 	0.7263 & 	0.7353  \\
        L-hook electroc.        & 	0.7450 & 0.5338    & 	0.6290 & 	0.6829 & 	0.6821  \\
        Liver                   & 	0.8000 & 0.7685    & 	0.7639 & 	0.8016 & 	0.7750  \\
        \hline
        Mean                    & 	0.6910 & 0.6110     & 	\textbf{0.6537} & 	0.6842 & 	\textbf{0.6938}  \\
        \specialrule{1.2pt}{0.2pt}{1pt}
    \end{tabular}
    }
\end{table}

\begin{table}[h]
    \centering
    \caption{Per-class TC for CholecSeg8k}\label{tab:ch-tc-per-class}
    \resizebox{\textwidth}{!}{%
    \begin{tabular}{lc|cc|cc}
        \specialrule{1.2pt}{0.2pt}{1pt}
        \textbf{Class} & \textbf{Mask2Former} & \textbf{HRNet32} & \multicolumn{1}{c}{\begin{tabular}[c]{@{}c@{}}\textbf{HRNet32} \\ \textbf{+SP-TCN}\end{tabular}} & \textbf{Swin Base} & \multicolumn{1}{c}{\begin{tabular}[c]{@{}c@{}}\textbf{Swin base} \\ \textbf{+SP-TCN}\end{tabular}}  \\
        \specialrule{1.2pt}{0.2pt}{1pt}
        Background              &	0.9900 & 0.9894 & 	0.9892 & 	0.9876 & 	0.9890  \\
        Abdominal wall          & 	0.9264 & 0.9268 & 	0.9110 & 	0.9110 & 	0.9337  \\
        Connective tissue       & 	0.7310 & 0.6241 & 	0.6870 &    0.6701 & 	0.7091   \\
        Fat                     & 	0.9600 & 0.9549 & 	0.9372 & 	0.9550 & 	0.9615  \\
        Gallbladder             &    0.8613 & 0.8820 & 	0.8974 & 	0.9130 & 	0.9203   \\
        Gastrointestinal tract  & 	0.7070 & 0.5880 & 	0.7944 & 	0.5733 & 	0.7182   \\
        Grasper                 & 	0.8422 & 0.8090 & 	0.8196 &    0.8283 & 	0.8508    \\
        L-hook electroc.        & 	0.8250 & 0.8245 & 	0.7889 & 	0.7891 & 	0.8164   \\
        Liver                    & 	0.9480 & 0.9321 & 	0.9359 & 	0.9370 & 	0.9546   \\
        \hline
        Mean                     & 	0.8659 & 0.8368 & 	0.8624 & 	0.8406 & 	0.8726   \\
        \specialrule{1.2pt}{0.2pt}{1pt}
    \end{tabular}
    }
\end{table}

\section{Conclusions}\label{sec:conclusions}
We present a temporal model that can be used with any segmentation encoder to transform it to a video semantic segmentation model. The model is based on the one-dimensional TCN model presented in~\cite{farha2019ms} and is modified to effectively use both spatial and temporal information. We validated its performance on two datasets, the public CholecSeg8k and the private PN dataset. Results showed that the proposed model consistently improves both segmentation and temporal consistency performance. We showed the feasibility of performing fine-grained semantic segmentation on PN, which has not been investigated before in the literature. Improving temporal consistency for models to be used during PN will facilitate correct identification of the renal vessels, and therefore support safe clamping of the renal artery to avoid significant blood loss during renal tumour removal. We also showed segmentation outputs for other anatomy present in the procedure, such as the kidney and the liver. Exploiting temporal information to make models more accurate is a step towards clinical use. 

\section*{Statements and Declarations}
\noindent\textbf{Conflict of Interests:} Drs. Grammatikopoulou, Sanchez-Matilla, Bragman, Owen, Luengo and Prof. Stoyanov are employees of Medtronic Digital Surgery. Prof. Stoyanov is a co-founder and share-
holder in Odin Vision, Ltd.

\noindent\textbf{Ethical approval:} Digital Surgery maintains all necessary rights and consents to process, analyze and display the private data referenced in this study.

\bibliography{sn-bibliography}


\end{document}